\documentclass{article}

\PassOptionsToPackage{numbers, compress}{natbib}
\usepackage[preprint]{neurips_data_2024}
\usepackage[utf8]{inputenc} % allow utf-8 input
\usepackage[T1]{fontenc}    % use 8-bit T1 fonts
\usepackage{hyperref}       % hyperlinks
\usepackage{url}            % simple URL typesetting
\usepackage{booktabs}       % professional-quality tables
\usepackage{amsfonts}       % blackboard math symbols
\usepackage{nicefrac}       % compact symbols for 1/2, etc.
\usepackage{microtype}      % microtypography
\usepackage[table,xcdraw]{xcolor}
\usepackage{graphicx}
\usepackage{booktabs}
\usepackage{colortbl}
\usepackage{multirow}
\usepackage[normalem]{ulem}
\usepackage{utfsym}
\usepackage{algorithm}
\usepackage{algorithmicx}
\usepackage{algpseudocode}
\usepackage{amsmath}
\usepackage{amssymb}
\usepackage[super]{nth}
\usepackage{xspace}
\usepackage{wrapfig}
\usepackage{fontawesome}
\usepackage{array}
\usepackage{enumitem}
\usepackage{bbding}
\usepackage[skip=1ex]{caption}
\usepackage{tabularx}

\definecolor{deeppurple}{RGB}{102, 0, 153}

\newcommand{\etal}{\textit{et al.}}
\newcommand{\ie}{\textit{i.e.}}

\title{MM-WLAuslan: Multi-View Multi-Modal Word-Level Australian Sign Language Recognition Dataset}

\author{%
Xin Shen \quad Heming Du \quad Hongwei Sheng \quad Shuyun Wang \quad Hui Chen\thanks{Work done while visiting the University of Queensland.} \quad Huiqiang Chen\footnotemark[1] \\ \quad \textbf{Zhuojie Wu} \quad \textbf{Xiaobiao Du\footnotemark[1]} \quad \textbf{Jiaying Ying} \quad \textbf{Ruihan Lu} \quad \textbf{Qingzheng Xu} \quad \textbf{Xin Yu}\thanks{Corresponding author.} \\
The University of Queensland \\
\texttt{x.shen3@uqconnect.edu.au}
}

\begin{document}
\renewcommand{\thefootnote}{\arabic{footnote}}
\maketitle
\begin{figure}[h]
\begin{center}
\vspace{-3em}
\includegraphics[width=\linewidth]{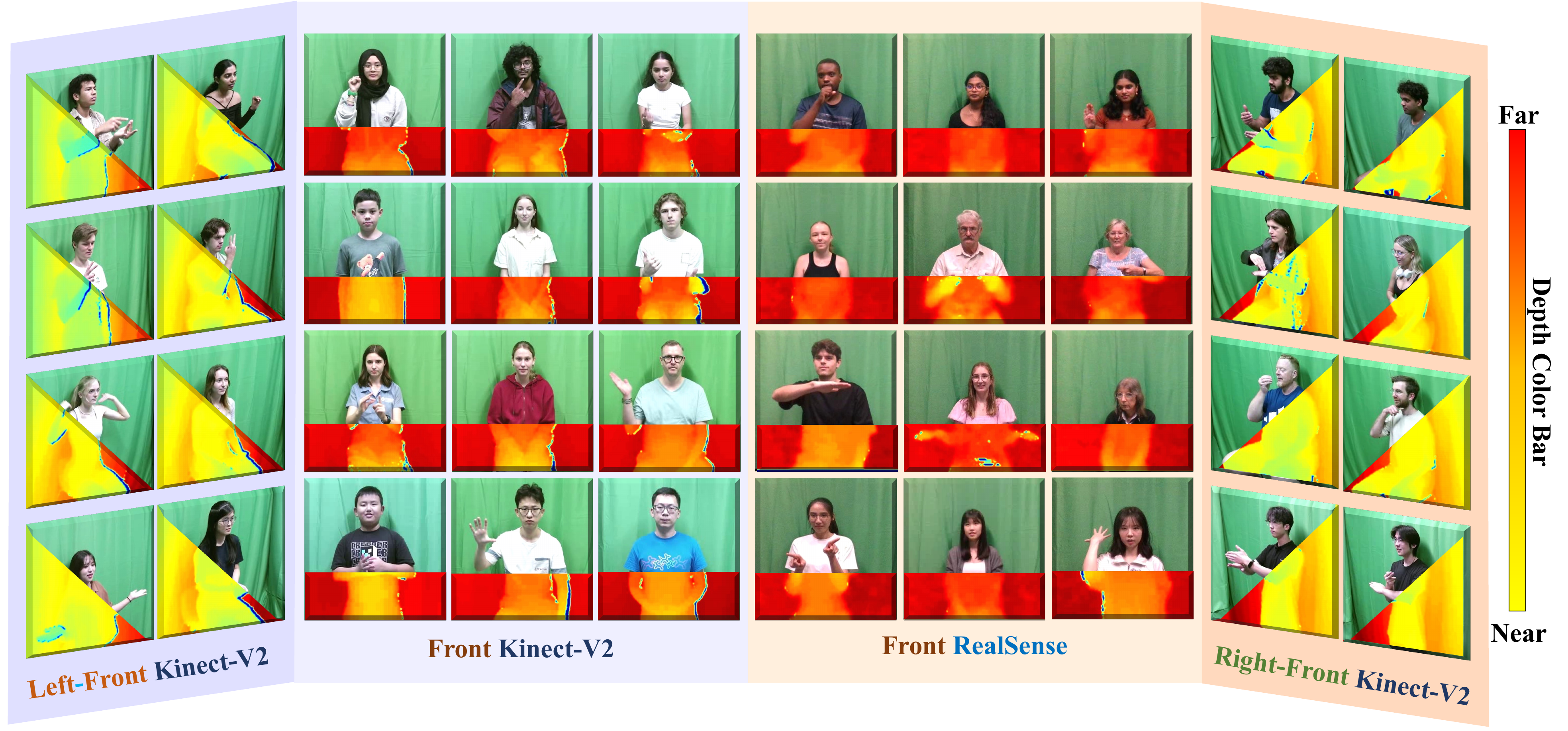}
\end{center}
\vspace{-1em}
\caption{\textbf{Illustrations of the curated MM-WLAuslan dataset.} MM-WLAuslan includes three Kinect-V2 cameras and a RealSense camera arranged hemispherically around the front half of the signer to capture multi-view and multi-modal data. 
% Additionally, MM-WLAuslan involves various signers from different cultural backgrounds.
}
\label{cases}
\vspace{-1em}
\end{figure} 

\begin{abstract}
% Considering different geographic regions generally have their own native sign languages, it is valuable to establish corresponding ISLR datasets to support related communication and research.
%, which includes over \textbf{3,200} commonly used Auslan glosses and more than \textbf{257k} sign videos recorded by \textbf{73} signers in a lab environment.
% Our filming system synchronizes three multi-view Kinect-V2 cameras and a centrally positioned RealSense camera.
% MM-WLAuslan has two main features: (1) the most extensive vocabulary, and (2) the recorded videos in our data are multi-view, multi-modal and high-quality.
% To mimic real-world scenarios, we conduct ISLR task in-domain and cross-domain settings for the .
% Specifically, the in-domain setting includes pixel-based, pose-based, and multi-modal-based ISLR within the same camera, whereas the cross-domain involves transfer learning across different camera types and perspectives.
Isolated Sign Language Recognition (ISLR) focuses on identifying individual sign language signs.
Considering the diversity of sign languages across geographical regions, developing region-specific ISLR datasets is crucial for supporting communication and research.
Auslan, as a sign language specific to Australia, still lacks a dedicated large-scale word-level dataset for the ISLR task.
To fill this gap, we curate \underline{\textbf{the first}} large-scale Multi-view Multi-modal Word-Level Australian Sign Language recognition dataset, dubbed MM-WLAuslan.
Compared to other publicly available datasets, MM-WLAuslan exhibits three significant advantages: (1) \underline{\textbf{the largest amount}} of data, (2) \underline{\textbf{the most extensive}} vocabulary, and (3) \underline{\textbf{the most diverse}} of multi-modal camera views. 
Specifically, we record \textbf{282K+} sign videos covering \textbf{3,215} commonly used Auslan glosses presented by \textbf{73} signers in a studio environment.
Moreover, our filming system includes two different types of cameras, \ie, three Kinect-V2 cameras and a RealSense camera. 
We position cameras hemispherically around the front half of the model and simultaneously record videos using all four cameras. 
Furthermore, we benchmark results with state-of-the-art methods for various multi-modal ISLR settings on MM-WLAuslan, including multi-view, cross-camera, and cross-view.
Experiment results indicate that MM-WLAuslan is a challenging ISLR dataset, and we hope this dataset will contribute to the development of Auslan and the advancement of sign languages worldwide.
All datasets and benchmarks are available at~\href{https://uq-cvlab.github.io/MM-WLAuslan-Dataset/}{\faGithub~\textcolor{deeppurple}{MM-WLAuslan}}.
% To mimic real-world scenarios, we report the performance under various settings, such as cross-domain, multi-model and cross-views. 
% that is trained and evaluated using the data from the same camera or across cameras. 
%in a broader context.

\end{abstract}

%=====================================
\section{Introduction}
Sign language (SL) is the primary mode of communication for many deaf or hard-of-hearing individuals. 
Each sign language possesses its own vocabulary and grammatical rules, akin to spoken languages~\cite{abner2020getting,sl_bg1,sl_bg3}. 
Notably, even within regions that share a commonly spoken language, such as the United States, Australia, and the United Kingdom, distinct native sign languages are prevalent. 
To facilitate communication between the deaf and hearing communities, Isolated Sign Language Recognition (ISLR) is highlighted as a fundamental sign language understanding task. 
ISLR aims to recognize an individual sign gloss, which is a written representation of signs using words from a spoken language, into a corresponding word or phrase in spoken languages~\cite{wlasl, auslan_daily}.
% \footnote{``Gloss'' refers to .} 
% Sign language (SL) is the primary way for deaf or people with hearing loss to express themselves. 
% Similar to various spoken languages, sign languages have their own vocabularies and grammar \cite{abner2020getting,sl_bg1,sl_bg3}.
% More importantly, diverse geographic regions usually have their native sign languages even though these regions share a commonly spoken language, such as America, Australia, and the UK.
% Isolated sign language recognition (ISLR) has been developed to bridge communication barriers between the deaf and hearing communities by converting an isolated sign gloss into a spoken language word or phrase~\cite{wlasl, auslan_daily}.

With emerging deep learning techniques~\cite{i3d,3dcnn,s3d,slowfast} and large-scale sign language datasets~\cite{wlasl,msasl,csl500,dk40,lsa64}, ISLR achieves promising progress recently~\cite{nla_slr,wlasl,SAM-SLR}.
As shown in Table~\ref{islr_datasets}, researchers from various countries construct word-level sign language datasets and thus promote the development of ISLR in the respective sign languages, such as American Sign Language (ASL) \cite{wlasl,Purdue_RVL-SLLL, ASLLVD, popsign_asl, ASL_Citizen, msasl}, British Sign Language (BSL) \cite{bsl-1k, bsldict}, Chinese Sign Language (CSL) \cite{CSL_Daily, DEVISIGN} and German Sign Language (DGS)~\cite{dk40, SMILE}. 
Meanwhile, according to the Australian Federal Department of Health and Aged Care (DHAC)\footnote{\url{https://www.health.gov.au/topics/ear-health/about}}, as of 14 May 2024, one in six Australians, over 3.6 million people, had hearing loss affecting them, and the number is expected to reach 7.8 million people by 2060.
However, to the best of our knowledge, there is no publicly available large-scale Auslan dataset for ISLR. 
Due to the regional nature of sign languages and the societal commitment to supporting individuals with hearing impairments, word-level Australian Sign Language (Auslan) datasets are inevitably and urgently needed in order to investigate automatic recognition.

Moreover, the volume of data, the breadth of data categories, and the diversity of data modalities are three critical points that influence the fundamental quality of an ISLR dataset. The larger the volume, the wider the range of categories, and the richer the modalities of data mean the higher the value of the dataset for scientific research and practical applications, such as sign language education~\cite{sheng2024ai} and dictionary~\cite{asl_learner}.
Specifically, a large volume of data and an extensive gloss dictionary within the dataset enhance the robustness and capability of the sign recognition system.
 % broadens the capabilities of the sign recognition system.
% A comprehensive dataset with an extensive gloss dictionary enables the recognition of a wider range of signs, improving the utility and robustness of the system. 
Additionally, the captured multi-view sign data and depth information improve the accuracy of recognizing complex hand movements and reduce the issues caused by occlusion and single-view ambiguities. 
% Incorporating depth information helps in capturing the three-dimensional aspects of sign language, such as hand shape and spatial orientation, which are essential for precise recognition. 
However, most existing publicly available ISLR corpora either contain the limited gloss videos and vocabulary~\cite{auslan_daily, Purdue_RVL-SLLL, RWTH-BOSTON-50, dk40, lsa64, SMILE, KL_MV2DSL} or are only captured in a single viewpoint without depth information~\cite{wlasl, msasl, ASL_Citizen, popsign_asl, bsl-1k}.

% Moreover, most existing publicly available ISLR corpora either contain the limited size of the gloss dictionary~\cite{auslan_daily, Purdue_RVL-SLLL, RWTH-BOSTON-50, dk40, lsa64, SMILE, KL_MV2DSL} or are only captured in a single viewpoint without depth information~\cite{wlasl, msasl, ASL_Citizen, popsign_asl, bsl-1k}.
% To address these limitations and enhance the practical applications~\cite{sheng2024ai,asl_learner} of ISLR systems, we aim to develop a dataset that not only encompasses a large gloss dictionary but also utilizes multi-view recording and incorporates depth cameras.
% Specifically, a comprehensive ISLR dataset with an extensive gloss dictionary enables the recognition of a wider range of signs, improving the utility and robustness of the system. 
% Multi-view sign data capture allows for the more accurate recognition of complex hand movements and reduces the issues caused by occlusion and single-view ambiguities. 
% Additionally, incorporating depth information helps in capturing the three-dimensional aspects of sign language, such as hand shape and spatial orientation, which are essential for precise recognition. 
% The integration of these features is pivotal in advancing the state-of-the-art in sign language recognition and enhancing the practical applications of these systems.

In this work, we record the first word-level Auslan recognition dataset, named MM-WLAuslan, that contains the largest number of data samples, the most extensive vocabulary, and the most diverse multi-modal camera views compared to other publicly available datasets, as shown in Table~\ref{islr_datasets}.
Specifically, we select 3,215 commonly used glosses that contain a sufficient variety of classes and training instances for a practical word-level Auslan recognition model. 
We collect the glosses from \textit{Auslan SignBank}\footnote{Auslan SignBank:~\url{https://auslan.org.au/dictionary/}}~\cite{Auslan_SignBank}, the most authoritative Auslan dictionary in Australia. 
We ask Auslan experts to help select glosses that are widely used throughout Australia, including fingerspelling glosses\footnote{English words are signed letter by letter.}, such as ``\textit{TV}'' and ``\textit{NEWS}''.
The collected glosses correspond to over 7,900 English words or phrases, covering most of the vocabulary commonly used in daily life. 
% select glosses that are used throughout Australia and fingerspelling\footnote{English words are signed letter by letter.} glosses that are recognized as commonly used by Auslan experts, such as ``\textit{TV}'' and ``\textit{NEWS}''.
We invite sign language experts, deaf individuals, and volunteers to participate in the recording process.
After 2,500+ hours of preparation and recording, we capture over 282K+ high-quality isolated Auslan gloss videos with the assistance of 73 signers.
Each video recording is supervised by at least one Auslan expert to ensure the precision of the sign language expression.
% To ensure the accuracy of the sign language videos, at least one sign language expert supervises the recordings. 
% If any signs are performed incorrectly, the expert points out the issues and the videos are re-recorded. 
% Most existing publicly available ISLR corpora either contain the limited size of the gloss dictionary~\cite{auslan_daily, Purdue_RVL-SLLL, RWTH-BOSTON-50, dk40, lsa64, SMILE, KL_MV2DSL} or are only captured in a single viewpoint without depth information~\cite{wlasl, msasl, ASL_Citizen, popsign_asl, bsl-1k}.
% To address these limitations, MM-WLAuslan not only encompasses a large gloss dictionary but also utilizes multi-view recording and incorporates depth cameras.
% Specifically, we collect 257k sign videos for 3,200 glosses to endow the ISLR systems trained on MM-WLAuslan with a robust ability of a wider range of ISLR.  %, improving the utility and robustness of the system. 
% Meanwhile, we capture multi-view sign videos to reduce the issues caused by occlusion and single-view ambiguities. 
% Additionally, incorporating depth information helps in capturing the three-dimensional aspects of sign language, such as hand shape and spatial orientation, which are essential for precise recognition. 
% The integration of these features is pivotal in advancing the state-of-the-art in ISLR and enhancing the practical applications of these systems.

To record multi-view, multi-modal, and high-quality isolated Auslan gloss videos, we set up a recording studio equipped with a green screen backdrop.  % capture in studio
We position two different types of RGB-D cameras, \ie, three Kinect-V2 cameras and a RealSense camera, hemispherically around the front half of the model. 
As shown in Figure~\ref{cases}, we place the cameras to the left-front, front, and right-front of the subject and simultaneously record videos.
Unlike the previous dataset~\cite{SMILE} that only provides depth video from the front view, we record both RGB-D videos from every camera. 
Furthermore, for an unbiased performance evaluation of ISLR systems, we involve nearly 20 signers in the test set who are not exposed to the training and validation sets. Concurrently, we split the test set into four distinct subsets to mimic the various scenarios in the real world. 
Videos in three subsets are designed to incorporate diverse backgrounds or potential temporal disturbances.
After obtaining the realistic test sets, we utilize the collected multi-modal, multi-view, and multi-camera videos to benchmark various multi-modal ISLR settings. Extensive experiments demonstrate the limitations of current state-of-the-art (SOTA) methods when these methods are applied across various cameras and views. This manifests the potential of MM-WLAuslan to advance the future research and development of ISLR systems. 
Overall, our major contributions are summarized as follows:
\begin{itemize}
    \item We construct the first word-level Australian ISLR dataset, dubbed MM-WLAuslan. MM-WLAuslan consists of the largest number of gloss videos and the most extensive vocabulary.
    \item We provide the most diverse multi-modal camera views and enable the investigation of a variety of multi-modal ISLR settings, including multi-view, cross-camera and cross-view.
    \item We establish a leaderboard and an evaluation benchmark to promote future Australian ISLR research and development of applications.
\end{itemize}

%=====================
\section{Related Work}

\begin{table*}[t]
\centering
\tiny
\caption{\textbf{Comparison between MM-WLAuslan and existing ISLR datasets.}}
\setlength{\tabcolsep}{1.25em}{
\begin{tabular}{m{2.5cm}ccccccccc}
\toprule
\textbf{Dataset} & \textbf{Country} & \textbf{Signs} & \textbf{Signers} & \textbf{Videos} & \textbf{Ave.Videos/Sign} & \textbf{Cross-Cam} & \textbf{Depth} & \textbf{Source}\\
\midrule
\rowcolor[HTML]{EFEFEF} Purdue RVL-SLLL~\cite{Purdue_RVL-SLLL} & USA & 39 & 14 & 0.5K & 14 & \usym{2717} & \Checkmark & Studio \\
\rowcolor[HTML]{EFEFEF} RWTH-BOSTON 50~\cite{RWTH-BOSTON-50} & USA & 50 & 3 & 0.5K & 9.66 & \Checkmark & \usym{2717} & Studio \\
\rowcolor[HTML]{EFEFEF} ASLLVD~\cite{ASLLVD} & USA & 3,000 & 6 & 9.8K & 3.27 & \Checkmark & \usym{2717} & Studio \\
\rowcolor[HTML]{EFEFEF} WLASL~\cite{wlasl} & USA & 2,000 & 119 & 21.1K & 10.54 & \usym{2717}& \usym{2717} & Web \\
\rowcolor[HTML]{EFEFEF} MS-ASL~\cite{msasl} & USA & 1,000 & 222 & 25.5K & 25.51 & \usym{2717} & \usym{2717} & Web \\
\rowcolor[HTML]{EFEFEF} ASL Citizen~\cite{ASL_Citizen} & USA & 2,731 & 52 & 83.9K & 30.73 & \usym{2717}& \usym{2717} & Webcam \\
\rowcolor[HTML]{EFEFEF} PopSign ASL v1.0~\cite{popsign_asl} & USA & 250 & 47 & 214.3K & 857.30 & \usym{2717} & \usym{2717} & Smartphone \\
% BSLDict~\cite{bsldict} & BSL & 9,283 & > 28 & 14.2k & 1.53 & 1 & \usym{2717} & Lab \\
BSL-1K~\cite{bsl-1k} & GBR & 1,064 & 40 & 273.0K & 257 & \usym{2717} & \usym{2717} & Web \\
% DEVISIGN-G~\cite{DEVISIGN} & CSL & 36 & 8 & 0.4k & 12.00 & 1& \Checkmark & Lab \\
% DEVISIGN-D~\cite{DEVISIGN} & CSL & 500 & 8 & 6.0k & 12.00 & 1& \Checkmark & Lab \\
\rowcolor[HTML]{EFEFEF} DEVISIGN-L~\cite{DEVISIGN} & CHN & 2,000 & 8 & 24.0K & 12.00 & \usym{2717} & \Checkmark & Studio \\
\rowcolor[HTML]{EFEFEF} CSL 500~\cite{csl500} & CHN & 500 & 50 & 125.0K & 250.00 & \usym{2717} & \Checkmark & Studio \\
DGS Kinect 40~\cite{dk40} & DEU & 40 & 14 & 2.8K & 70.00 & \usym{2717} & \Checkmark & Studio \\
\rowcolor[HTML]{EFEFEF} SMILE~\cite{SMILE} & DEU/CHE & 100 & 30 & - & - & \Checkmark & \Checkmark & Studio \\
GSL 982~\cite{GSL_982} & GRC & 982 & 1 & 4.9K & 5.00 & \usym{2717} & \usym{2717} & Studio \\
\rowcolor[HTML]{EFEFEF} INCLUDE~\cite{INCLUDE} & ISR & 263 & 7 & 4.3K & 16.30 & \usym{2717} & \usym{2717} & Studio \\
\rowcolor[HTML]{EFEFEF} KL-MV2DSL~\cite{KL_MV2DSL} & ISR & 200 & - & 5.0K & 25 & \Checkmark & \usym{2717} & Studio\\
LSA64~\cite{lsa64} & ARG & 64 & 10 & 3.2K & 50.00 & \usym{2717} & \usym{2717} &Studio \\
\rowcolor[HTML]{EFEFEF} LSE-Sign~\cite{LSE-sign} & ESP & 2,400 & 2 & 2.4K & 1.00 & \Checkmark & \usym{2717} &Studio \\
LSFB-ISOL~\cite{LSFB-ISOL} & FRA/BEL & 395 & 100 & 47.6K & 120.38 & \usym{2717} & \usym{2717} & Studio \\
\rowcolor[HTML]{EFEFEF} BosphorusSign22K~\cite{BosphorusSign22K} & TUR & 744 & 6 & 22.5K & 30.30 & \usym{2717}& \Checkmark & Studio \\
\rowcolor[HTML]{EFEFEF} AUTSL~\cite{AUTSL} & TUR & 226 & 43 & 38.3K & 169.63 & \usym{2717} & \Checkmark & Studio \\
\midrule
% Auslan SignBank~\cite{Auslan_SignBank} & Auslan & 5,000 & - & 5k & 1.00 & 1& \usym{2717} & Lab \\
Auslan-Daily~\cite{auslan_daily} & AUS & 600 & 21 & 3.0K & 5.00 & \usym{2717} & \usym{2717} & Web  \\ 
\rowcolor[HTML]{EFEFEF}\textbf{MM-WLAuslan} & \textbf{AUS} & \textbf{3,215} & \textbf{73} & \textbf{282.9K} & \textbf{88.00} & \CheckmarkBold & \CheckmarkBold& \textbf{Studio} \\
\bottomrule
\end{tabular}
}
\label{islr_datasets}
\end{table*}

% \subsection{Isolated Sign Language Recognition Datasets}
\subsection{Isolated Sign Language Recognition Datasets}
As shown in Table~\ref{islr_datasets}, several datasets are developed to facilitate research and application development of ISLR.
However, most datasets have limitations in gloss dictionary size, depth information, and recording perspectives. 
For example, Purdue RVL-SLLL dataset~\cite{Purdue_RVL-SLLL} exhibits methodological rigor in a laboratory setting, but its applicability for sign language recognition is limited because it only covers 39 signs.
% the Purdue RVL-SLLL dataset~\cite{Purdue_RVL-SLLL}, despite its methodological rigour in a laboratory setting, covers only 39 signs, which substantially limits its applicability for comprehensive sign language recognition. 
% Furthermore, although ASLLVD dataset~\cite{ASLLVD} includes a large lexicon of 3,000 glosses, it only offers videos from a single perspective and lacks depth information, which is essential for capturing the three-dimensional motion of sign language.
Furthermore, despite ASLLVD dataset~\cite{ASLLVD} including a large lexicon of 3,000 glosses, it is limited by its single perspective and lack of depth information, crucial for capturing the three-dimensional motion of sign language.
WLASL~\cite{wlasl} and MS-ASL~\cite{msasl} datasets expand on the number of signs and signers but still restrict their recordings to single-view videos without depth, missing critical spatial dynamics essential for accurate sign interpretation. 
In contrast, datasets like CSL 500~\cite{csl500} and DGS Kinect 40~\cite{dk40} include depth information but cover only a small number of glosses, limiting their usefulness for extensive sign language applications.
% These datasets are collected in controlled lab environments, which improves data accuracy and consistency. 
% However, they have limited sign diversity and only single-view recordings. 

Different from all of the above datasets, the proposed MM-WLAuslan dataset is a comprehensive ISLR dataset.
It encompasses 3,215 signs from 73 signers, with each sign captured from four distinct viewpoints along with depth information, significantly enhancing the diversity and utility of the dataset. 
% This comprehensive multi-dimensional capture approach provides a robust foundation for the development of more accurate and flexible sign language recognition systems.
Moreover, MM-WLAuslan is currently the largest sign language recognition dataset in Australia, with extensive lexicon and high-quality data. 

\subsection{Isolated Sign Language Recognition Methods}
ISLR aims to identify the gloss labels of short-term videos.% ~\cite{3dposeislr}
Previous research can be categorized into three types based on the input modality: pixel-based, pose-based and multi-modal-based approaches. \\
% ISLR task is to classify a short-term video by its corresponding gloss label. 
% Research in ISLR primarily falls into three categories based on the input modality: pixel-based, pose-based and multi-modal-based approaches. \\
\textbf{Pixel-based ISLR: }Significant advances in CNN-based action recognition inspire the development of pixel-based ISLR models. 
Early efforts~\cite{cnn1,cnn2,cnn3} utilize convolutional neural networks (CNN) to extract frame-wise features, which are then temporally encoded using recurrent neural networks to capture time-series information. 
% Similar to those used in action recognition, are widely adopted. 
Meanwhile, 3D CNNs, such as C3D model~\cite{3dcnn1,3dcnn2,3dcnn3} and I3D model~\cite{i3d}, are commonly used in ISLR~\cite{msasl, wlasl, i3d_pose, ASL_Citizen, our_spotting, transfer_islr}. \\
\textbf{Pose-based ISLR:} Unlike RGB pixel-based methods, pose-based ISLR models are robust against background clutters, lighting conditions, and occlusions, while explicitly depicting human hand and limb movements~\cite{pose1,pose2,pose3, pose_ar}.
ST-GCN~\cite{pose3}, the first to apply a spatio-temporal graph convolutional network for action recognition, encodes motions across the human kinetic chain. 
Subsequent studies utilize this spatio-temporal architecture, employing both graph convolutional networks \cite{wlasl,pose_islr_rw_1, DSTA-SLR, STC-SLR} and Transformers~\cite{vaswani2017attention, pose_islr_rw_2, pose_islr_rw_3, 3dposeislr} to embedding and analyze sign pose data. \\
\textbf{Multi-modal-based ISLR: }Recent studies show that combining pose, depth, and RGB modalities significantly improves ISLR. 
Zuo et al.~\cite{nla_slr} use the S3D model to extract RGB and pose heatmap features, enhancing recognition on the WLASL~\cite{wlasl} dataset. 
Moreover, Jiang et al.~\cite{SAM-SLR} integrate depth information into the model, enabling recognition results to exceed 99\% on the AUTSL dataset~\cite{AUTSL}.

\subsection{Multi-view and Multi-modal Action Recognition}

Previous research~\cite{sarhan2020transfer} argues that Action Recognition (AR) methods can be applied on sign language recognition. 
To build an effective and robust real-world ISLR and AR system, initiating multi-view and multi-modal learning is essential~\cite{KL_MV2DSL}. 
Recent advancements in AR introduce various approaches for multi-view learning~\cite{multi_view_model}, including dictionary learning~\cite{mv_1}, neural networks with adjustable views~\cite{mv_2}, convolutional neural networks~\cite{mv_3}, and attention mechanisms~\cite{mv_4}. 
% Deep learning networks emerge as the most researched and recognized models in this area. 
Additionally, Zhu~\etal~\cite{mv_5} adopt vision transformer models as robust solutions for multi-view learning. 
% These models integrate view-specific features, which are then processed by dense layers to enhance classification accuracy.
% Simultaneously, multi-modal action recognition leverages data from diverse sources to improve performance by utilizing the unique motion information from different modalities. 
Recent approaches~\cite{mm_1, mm_2} develop robust view-invariant representations for downstream tasks, while DA-Net~\cite{mm_3} merges view-specific and independent modules for effective prediction. 
A feature factorization approach in~\cite{mm_4} and a cascaded residual autoencoder in~\cite{mm_5} address challenges in RGB-D action recognition and incomplete view classification, respectively. 

\section{Proposed MM-WLAuslan Dataset}
\label{sec:dataset}

In this section, we describe our recording setup and workflow, detail the data processing and augmentation, and provide statistics for the MM-WLAuslan\footnote{Our dataset follows the copyright~\textbf{Creative Commons BY-NC-SA 4.0} license~\href{https://creativecommons.org/licenses/by-nc-sa/4.0/} {\faCopyright}. Please note that we obtain the consent of the signers before recording them.} dataset. 

\subsection{Recording Setup and Workflow}
Our recording setup is located in a studio environment surrounded by a green screen. 
% This not only provides a clean background but also facilitates the replacement of different real-world backgrounds. 
In the studio, we position Kinect-V2 cameras at the left-front, front, and right-front views, along with a centrly placed RealSense camera. 
Both Kinect-V2 and RealSense are capable of recording high-quality videos with depth information. 
In the Appendix, we compare the different parameters of these two types of cameras. 
Most importantly, the imaging principles of Kinect-V2 and RealSense cameras are different. 
The former employs time-of-flight technology to measure depth, while the latter utilizes stereo vision to capture depth information.
Moreover, Kinect-V2 offers high resolution and excellent depth sensing, while RealSense provides a higher frame rate and portability. 
We record data using these two types of RGB-D cameras to investigate the cross-camera robustness of methods.
% select the appropriate camera data for training the ISLR model based on specific needs and scenarios.

We recruit signers with diverse experience in Auslan, including Auslan experts, deaf individuals who use Auslan, and volunteers interested in sign language, to sign glosses\footnote{\textbf{\textit{Auslan experts}} refers to non-deaf individuals who are proficient in Australian Sign Language, while \textbf{\textit{non-expert deaf signers}} only refers to deaf individuals who use Auslan.}. 
The involvement of Auslan experts and deaf individuals ensures the precision of a subset of the data, which is crucial for precise research and applications of sign language. 
The extensive participation of volunteers enhances the diversity of the signers, reflecting the natural variability in the deaf community. 
% Meanwhile, we raise awareness about the deaf community, advocating for greater support for this marginalized group.
Moreover, we design an interactive interface for dataset recording and present the interface in the Appendix. 
% The first interface synchronizes video recording from four cameras. 
% It includes basic functions such as start, stop, pause, and input information. 
% The second interface is for sign language learning. 
% It plays commonly used Auslan glosses from Auslan SignBank to volunteers who do not know Auslan. 
We record videos of sign language imitated by volunteers. Each sign is supervised and checked by at least one expert to ensure the precision of the sign language expression.
% Volunteers learn and imitate the signs, and their performances are recorded.
% At least one sign language expert supervises the recordings to ensure the accuracy of the sign language videos. 
% If any signs are performed incorrectly, the expert points out the issues, and the videos are re-recorded. 

\begin{figure}[t]
\begin{center}
    \includegraphics[width=1\linewidth,]{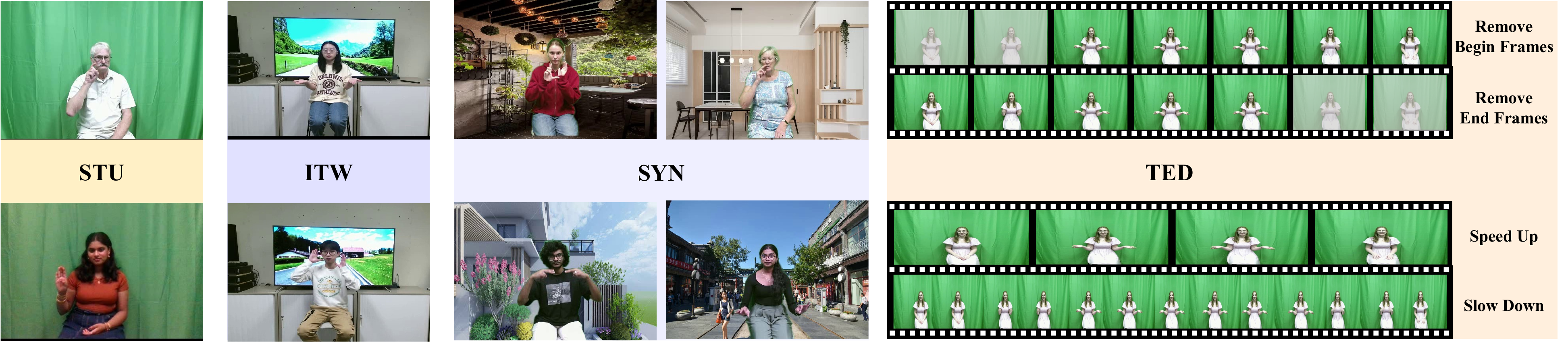}
    \caption{\textbf{Demonstrations of test subsets.} ``STU'', ``ITW'', ``SYN'', and ``TED'' represent the studio set, in-the-wild set, synthetic background set and temporal disturbance set, respectively.}
   % \caption{\textbf{Demonstrations of frame examples.} (a) Example frames from different cameras (Front Kinect-V2, Left-Front Kinect-V2, Right-Front Kinect-V2, Front RealSense) showing depth, RGB, and pose from left to right in order. 
   % (b) Example frames from indoor and outdoor scenes to demonstrate data diversity.}
\label{data_samples}
\end{center}
\vspace{-2em}
\end{figure}

\subsection{Data Processing and Augmentation}
After recording all the sign language videos, we notice that a significant portion of the footage consists of a green screen background.
% the original video resolution is too high for efficient storage, and 
Therefore, we utilize the keypoints estimated by AlphaPose~\cite{alphapose, fang2017rmpe,li2019crowdpose} to remove the background that is irrelevant to the sign language.
We crop videos based on a fixed-size box that can cover every signer and align their eyes on the same horizontal level.
% across all the data and then align the eyes on the same horizontal level based on the
% AlphaPose is renowned for its excellent human track and keypoint estimation capabilities. 
% By identifying the largest bounding box across all the data, we crop each dataset to focus solely on the individuals. 
% Simultaneously, we crop the corresponding depth for the original depth images.

% Due to our data being recorded in a clean laboratory environment, it lacks the variability present in real-world scenarios.
To evaluate the performance of ISLR systems under real-world scenarios, we provide a diverse test set with four distinct subsets, including studio (STU) set, in-the-wild (ITW) set, synthetic background (SYN) set, and temporal disturbance (TED) set. Each subset encompasses videos for all gloss vocabulary.
The \textbf{STU set} includes consistent scene settings with the training set.
In the \textbf{ITW set}, green screens are removed and replaced with dynamic or static backgrounds to simulate videos recorded in diverse environments, as shown in Figure~\ref{data_samples}.
We utilize the Background Remover\footnote{\url{https://github.com/nadermx/backgroundremover}} to extract signers from videos and synthesize indoor and outdoor backgrounds in the \textbf{SYN set}.
The \textbf{TED set} simulates potential recording time discrepancies in real-world scenarios by randomly adjusting video segments through removal or altering playback speed.
% Therefore, we hypothesize that sign language in the real world would be subject to background interference and temporal disturbance. 
% Based on these hypotheses, we modify the partly test data.
% To simulate background interference, we first select 10 indoor scenes and 10 outdoor scenes. 
% We then use the Background Remover\footnote{\url{https://github.com/nadermx/backgroundremover}} to randomly replace the backgrounds in the test data. 
% In Figure~\ref{data_samples}(b), we present samples from the test set with replaced backgrounds.
% To simulate temporal disturbance, we assume that sign language videos may be incomplete or have variable motion speeds in real life. 
% Consequently, for the former, we randomly delete either the first or last quarter of the video, and for the latter, we apply random speed variations, including both acceleration and deceleration.

Overall, each data sample in our dataset includes: (1) RGB-D videos captured by a Kinect-V2 camera or a RealSense camera; (2) intrinsic and extrinsic parameters for the captured camera; (3) pose sequences corresponding to the RGB video; (4) gloss identities; (5) spoken English words or phrases corresponding to the gloss and (6) signer identities.
These various views and modalities of sign language video samples can be further investigated for different word-level Auslan ISLR settings.

\subsection{Data Statistics}
% & \usym{2717} & \Checkmark
\begin{wrapfigure}{r}{0.63\textwidth} % 表格将放在右侧，占据半栏宽度
    \vspace{-1.25em} % 调整表格与上方文字的间距
    \begin{minipage}{0.60\textwidth}
        \scriptsize
        \centering
        \captionof{table}{\textbf{Key statistics of MM-WLAuslan dataset splits.} 
        ``BG'' and ``TP'' represent background and temporal, respectively.
        % ``STU'', ``ITW'', ``SYN'', ``TED'', ``BG'', and ``TP'' represent the studio, in the wild, synthetic, temporal disturbance, background and temporal, respectively.
        ``OOS'' indicates the signers only occur in the test set.}
        \label{statistics}
        \setlength{\tabcolsep}{0.65em}{
        \begin{tabular}{l|c|c|c@{\hskip 0.05in}c@{\hskip 0.05in}c@{\hskip 0.05in}c}
            \toprule
            Split      & Train & Val & Test-STU & Test-ITW & Test-SYN  & Test-TED    \\ \midrule
            Num. Videos & 154.3k &25.7k &25.7k &25.7k &25.7k &25.7k \\
            Num. Signers & 55 & 53 & 12& 15& 62 & 63 \\
            Num. OOS & - & - & 10 & 2 & 15 & 10 \\
            BG Interference & \usym{2717} & \usym{2717} & \usym{2717} & \Checkmark & \Checkmark & \usym{2717} \\ 
            TP Disturbance  & \usym{2717} & \usym{2717} & \usym{2717} & \usym{2717} & \usym{2717} & \Checkmark \\ 
            \bottomrule
        \end{tabular}
        }
        \vspace{-1em} % 调整表格与下方文字的间距
    \end{minipage}
\end{wrapfigure}
\begin{figure}[t]
\begin{center}
    \includegraphics[width=1\linewidth,]{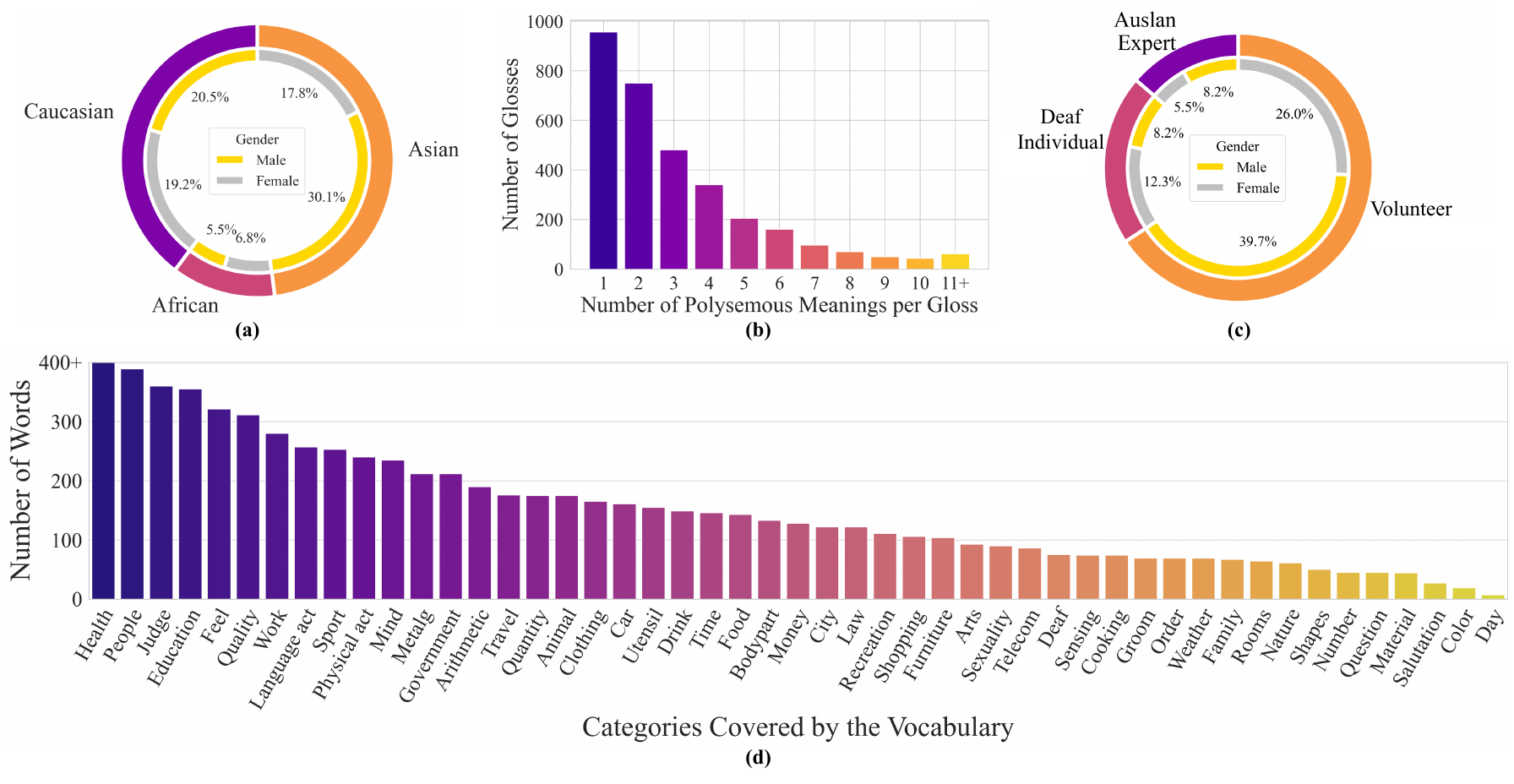}
   \caption{\textbf{Statistics of signers and glosses.} (a) Ethnicity and gender distribution. (b) Frequency of polysemous glosses. (c) Distribution of Auslan proficiency. (d) Categories of words.}
\label{stat}
\end{center}
\vspace{-2em}
\end{figure}

We select 3,215 commonly used Auslan glosses, corresponding to over 7,900 English words or phrases. 
As illustrated in Figure~\ref{stat}(b), there are more than 2,000 glosses with multiple meanings, highlighting the contextual variability of sign language similar to natural languages.
Additionally, these terms are finely categorized into 49 groups, including health, education, and others, as shown in Figure~\ref{stat}(d).
The extensive vocabulary and semantic richness of MM-WLAuslan demonstrate its potential to advance sign language research and applications.

After over 2,500 hours of recording, we capture 282,900 videos by 73 signers. 
Specifically, for 3,215 commonly used word-level Auslan glosses, we record every gloss 22 times utilizing 4 different cameras (3215×20×4). 
Unlike other datasets~\cite{wlasl, msasl}, our dataset maintains a consistent number of videos per Auslan gloss, thereby establishing a uniform ISLR dataset.
We split the samples of a gloss into training, validation, and testing sets following a ratio of 6:1:4. 
% Note that there are 18 signers that only exist in the testing set instead of overlapping among signers in the training and validation sets.
Note that the test set contains 18 signers who do not appear in either the training or validation sets.
Additionally, we further divide the testing set into the STU set, the ITW set, the SYN set, and the TED set in a 1:1:1:1 ratio. 
The detailed split statistics are demonstrated in Table~\ref{statistics}.

% To construct our dataset, we recruit volunteers with varying levels of proficiency in Auslan, including Auslan experts, deaf individuals who use Auslan, and Auslan learners. 
% The involvement of Auslan experts and deaf participants ensures the accuracy and professionalism of a subset of the data, which is crucial for precise research and application of sign language. 
% The extensive participation of learners enhances the diversity of the data, reflecting the natural variability and wide usage of sign language expressions. 
Moreover, as illustrated in Figure~\ref{stat}(a), we provide the ethnic and gender distribution of signers in MM-WLAuslan. The signers are categorized into three primary ethnic groups: Caucasian, African, and Asian. The male-to-female ratios are relatively balanced across the different ethnic groups.
% the diversity in racial composition among the contributors to our Auslan dataset is crucial for representing the multicultural context of Australia. 
% Our dataset includes participants from three primary racial groups: Caucasian, African, and Asian. 
% The ethnic diversity and the near-equitable gender balance within each ethnic group enrich the dataset. Consequently, the ISLR models trained on MM-WLAuslan are likely to be more effective and widely applicable across the diverse Australian population.
The near-equitable gender balance within each ethnic group not only enhances the representativeness of the dataset but also underscores its gender fairness.
Meanwhile, we include a broader range of ethnicities to enhance the inclusivity and representativeness of the dataset further. Thus, this composition ensures that the ISLR models developed from this dataset mitigate biases and offer equitable performance across the diverse Australian population.
Furthermore, as shown in Figure~\ref{stat}(c), we demonstrate the distribution of participants involved in recording, segmented by their proficiency in Auslan. 
We make concerted efforts to include as many Auslan experts and deaf individuals as possible for the quality of the recordings. Additionally, we recruit many volunteers to further increase the diversity of the signers, and thus, enrich the representativeness of the dataset. 

%=====================================
\section{MM-WLAuslan Benchmark}
In this section, we present and analyze benchmark results of various multi-modal ISLR settings on MM-WLAuslan.
% , including single-view RGB-based, single-view RGB-D-based, multi-view RGB-based, multi-view RGB-D-based ISLR, cross-camera and cross-view ISLR.
More experiments and details are included in the Appendix.

\begin{table}[t]
\centering
\tiny
\caption{\textbf{The baseline of Single-view RGB-based ISLR on MM-WLAuslan.} 
``STU'', ``ITW'', ``SYN'', ``TED'', and ``AVG.'' represent the studio set, in-the-wild set, synthetic background set, temporal disturbance set and average performance across the four subsets, respectively.
\textbf{Bold} indicates the highest value within the same data type.}
\label{svrgb}
\setlength{\tabcolsep}{1.15em}{
\begin{tabular}{l|c|cc|cc|cc|cc|cc}
\toprule 
\multirow{2}{*}{Model} & \multirow{2}{*}{Data Type} & \multicolumn{2}{c|}{\textbf{STU}} &  \multicolumn{2}{c|}{\textbf{ITW}} & \multicolumn{2}{c|}{\textbf{SYN}} & \multicolumn{2}{c|}{\textbf{TED}} & \multicolumn{2}{c}{\textbf{AVG.}} \\
 & & Top-1 & Top-5  & Top-1 & Top-5 & Top-1 & Top-5 & Top-1 & Top-5 & Top-1 & Top-5  \\ \midrule \midrule
ResNet2+1D~\cite{3dcnn} & Pixel & 58.71&77.03 & 13.83&18.37 & 26.14&39.58 & 51.14&69.97 & 37.45&51.24  \\
TSN~\cite{tsn} &  Pixel &51.17&68.60&11.06&23.75&31.01&45.89&40.40&69.10&33.41&51.84 \\
I3D~\cite{i3d} & Pixel & 63.97 &  84.93 & 14.18& 26.52 & 36.17 & 57.22 & 60.96 & 80.63 & 43.82 & 62.33\\
S3D~\cite{s3d} &  Pixel &75.55 & 94.11 & 29.41 & 55.11 & 44.60 &  71.34 &62.21 &85.26 & 52.94 & 76.46    \\
SlowFast~\cite{slowfast} &  Pixel & 80.68 & 96.08 &32.22& 64.81&53.17 & 78.30  & 66.21 & 82.18 &58.07& 80.34   \\
Timesformer~\cite{timesformer} &  Pixel &73.20&81.40&21.14&56.44&41.88&65.83&68.40&79.67&51.15&70.84    \\
UMDR~\cite{UMDR} &  Pixel &80.86&	95.88&	13.57&	28.66&	13.99&	31.01&	\textbf{82.69}&	\textbf{95.67}&	47.78&	62.81\\
KVNet-V~\cite{nla_slr} & Pixel &\textbf{84.51}&	\textbf{97.57}&	\textbf{39.88}&	\textbf{68.00}&	\textbf{56.56}&	\textbf{82.18}&	70.31&	90.86&	\textbf{62.82}&	\textbf{84.65}  \\
\midrule
TGCN~\cite{wlasl} & 2D pose & 68.62 & 86.30 & 58.01 & 74.74&  63.50 & 81.38 & 47.68 & 68.82 &62.11 & 77.81   \\
SL-GCN~\cite{SAM-SLR} & 2D pose & 71.07&91.21&66.59&89.5&63.20&86.94&\textbf{69.98}&88.99&67.71&89.16      \\ 
% ST-GCN~\cite{pose3} & 2D pose &  &  &  &  &   & \\
SPOTER~\cite{SPOTER} & 2D pose & 72.81 & 92.69 &64.12 & 86.36 & 66.81 & 88.11 & 69.42 & \textbf{90.94} &68.29& 89.53 \\
DSTA-SLR~\cite{DSTA-SLR} & 2D pose & 82.33 &	96.31 &	74.96 &	93.98 &	78.10 &	93.78 &	66.84 &	88.99 &	75.55 &	93.26 \\
STC-SLR~\cite{STC-SLR} & 2D pose & 79.92 &	95.88 &	74.35 &	93.92 &	76.02 &	93.50 &	63.11 &	87.33 &	73.35 &	92.65 \\
KVNet-K~\cite{nla_slr} & 2D pose & \textbf{82.88} &	\textbf{96.70} &	\textbf{76.29} &	\textbf{94.56} &	\textbf{79.07} &	\textbf{94.07} &	69.05 &	89.80 &	\textbf{76.82} &	\textbf{93.78} \\
\midrule
SAM-SLR~\cite{SAM-SLR} & 2D pose + Pixel & 83.98 & 97.12  & 74.30 & 91.65 & 80.73 & 94.93 &71.21 &86.56 &77.55 & 83.91\\
NLA-SLR~\cite{nla_slr} & 2D pose + Pixel & \textbf{86.32} &	\textbf{97.79} &	\textbf{79.05} &	\textbf{94.91} &	\textbf{84.26} &	\textbf{96.16} &	\textbf{77.98} &	\textbf{91.76} &	\textbf{81.90} &	\textbf{95.16}
\\
\bottomrule
\end{tabular}}
\end{table}

\begin{table}[t]
\centering
\tiny
\caption{\textbf{The baseline of Single-view RGB-D-based ISLR on MM-WLAuslan.} }
\label{svrgbd}
\setlength{\tabcolsep}{1.05em}{
\begin{tabular}{l|c|cc|cc|cc|cc|cc}
\toprule 
\multirow{2}{*}{Model} & \multirow{2}{*}{Data Type} & \multicolumn{2}{c|}{\textbf{STU}} &  \multicolumn{2}{c|}{\textbf{ITW}} & \multicolumn{2}{c|}{\textbf{SYN}} & \multicolumn{2}{c|}{\textbf{TED}} & \multicolumn{2}{c}{\textbf{AVG.}} \\
   &  & Top-1 & Top-5  & Top-1 & Top-5 & Top-1 & Top-5 & Top-1 & Top-5 & Top-1 & Top-5  \\ \midrule \midrule
I3D~\cite{i3d} & Pixel + Depth &65.74&88.57&21.71&41.32&61.06&85.41&47.25&65.71&48.94&70.25     \\
S3D~\cite{s3d} &  Pixel + Depth & 79.70& 95.93  & 64.97&89.16 & 76.38 &92.67 &66.11& 88.62 &71.79&91.60      \\
KVNet-V~\cite{nla_slr}&  Pixel + Depth &82.22 &96.75  & 38.79 &66.11  & 57.88 & 82.92 &66.94&88.58& 61.46& 83.59 \\
UMDR~\cite{UMDR}  &  Pixel + Depth& \textbf{91.65}&	\textbf{98.81}&	\textbf{72.52}&	\textbf{90.46}&	\textbf{83.77}&	\textbf{95.18}&	\textbf{88.35}&	\textbf{98.07}&	\textbf{84.07}&	\textbf{95.63}  \\
\midrule
TGCN~\cite{wlasl} & 3D pose & 70.19&89.78&59.52&76.59&66.35&84.06&51.48&71.17&61.88&80.40        \\
% ST-GCN~\cite{pose3} & 3D pose &  &  &  &  &  &   \\
SPOTER~\cite{SPOTER} & 3D pose & 74.95&95.88&66.75&89.41&70.22&91.23&71.65&92.36&70.89&92.22 \\
SL-GCN~\cite{SAM-SLR} & 3D pose &\textbf{77.76}&\textbf{96.98}&\textbf{72.26}&\textbf{91.49}&\textbf{74.91}&\textbf{92.57}&\textbf{72.27}&\textbf{94.88}&\textbf{74.30}&\textbf{93.98}   
\\ \midrule
NLA-SLR~\cite{nla_slr} & 2D pose + Pixel + Depth & 85.65&95.65&80.20&95.58&\textbf{83.36}&94.04&83.34&\textbf{94.63}&83.14&94.98   \\
SAM-SLR~\cite{SAM-SLR} & 3D pose + Pixel + Depth & \textbf{87.05}&\textbf{98.93}&\textbf{81.29}&\textbf{96.92}&83.03&\textbf{95.86}&\textbf{85.07}&93.53&\textbf{84.11}&\textbf{96.31}   \\
\bottomrule
\end{tabular}}
\end{table}

\subsection{Isolated Sign Language Recognition Settings}
\textbf{Single-view RGB-based ISLR} involves recognizing isolated sign language from video sequences captured from a single fixed camera. The input consists of RGB frames, denoted as $\{F_{1}, F_{2}, \ldots, F_{n}\}$, where $n$ represents the total number of frames in a video sequence. The single-view RGB setting utilizes spatial and temporal information from a singular perspective.

\textbf{Single-view RGB-D-based ISLR} aims to enhance the recognition of isolated signs by incorporating depth information along with RGB data. The input is represented as $\{(F_{1}, D_{1}), (F_{2}, D_{2}),$ $\ldots, (F_{n}, D_{n})\}$, where $D_{i}$ indicates the depth information corresponding to the $i$-th frame. This approach facilitates a richer interpretation of spatial dynamics. 

\textbf{Multi-view RGB-based ISLR} employs multiple cameras to capture the sign language videos. The input from each camera $k$ is represented as a sequence of RGB frames $\{F^{k}_{1}, F^{k}_{2}, \ldots, F^{k}_{n}\}$. 
Multi-view RGB data helps in mitigating issues related to occlusions and varied angles. 

\textbf{Multi-view RGB-D-based ISLR} incorporates depth data in a multi-view setup, the input for each camera $k$ is represented as $\{(F^{k}_{1}, D^{k}_{1}), (F^{k}_{2}, D^{k}_{2}), \ldots, (F^{k}_{n}, D^{k}_{n})\}$. 
This method enhances the model’s capability to interpret complex gestures from multiple perspectives. 

\textbf{Cross-Camera ISLR} aims to test the robustness of the model against variations in camera specifications and settings. Training and testing data are captured from different cameras.
It is challenging for the model to generalize across hardware-induced discrepancies. 

\textbf{Cross-View ISLR} requires the model to recognize signs from views not seen during training. With training views denoted as $V_{\text{train}}$ and testing views as $V_{\text{test}}$, the model must handle the appearance changes due to different viewing angles, thus testing its view-invariance capabilities.

\subsection{Evaluation Metric}
\textbf{Top-\textit{k} Accuracy} is quantitatively defined as the proportion of test instances for which the true label is among the top \textit{k} predictions made by the model. 
It is particularly suitable for ISLR~\cite{wlasl, msasl, auslan_daily} task with a large set of possible outcomes.
The expression is shown by the following equation:

\begin{equation}
\text{Top-}k \ \text{Accuracy} = \frac{1}{N} \sum_{i=1}^{N} \mathbf{1}(y_i \in \hat{Y}_i^k),
\end{equation}
where $N$ is the total number of instances in the test set, $\mathbf{1}$ is a binary indicator that returns 1 if the true label of the \(i\)-th instance $y_i$ is within the set of the top-$k$ predicted labels $\hat{Y}_i^k$ for that instance.
% and 0 otherwise.

% \begin{equation}
% \text{Top-\textit{k} Accuracy} = \frac{1}{N} \sum_{i=1}^{N} \mathbf{1}(\text{true label}_i \in \text{set of top-k predicted labels}_i),
% \end{equation}
% where $N$ is the total number of instances in the test set and $\mathbf{1}(\cdot)$ is an indicator function that returns 1 if the true label of the \(i\)-th instance, $\text{true label}_i$, is within the set of the top k predicted labels for that instance, $\text{set of top-k predicted labels}_i$, and 0 otherwise.
% This metric provides a useful measure of a model's predictive accuracy by considering a range of possible correct predictions rather than a strict exact match, thus accommodating the inherent uncertainty and variability in complex classification scenarios.

\subsection{Benchmark Results}

\begin{table}[t]
\centering
\tiny
\caption{\textbf{The baseline of Multi-view RGB-based ISLR on MM-WLAuslan. }``STU'', ``ITW'', ``SYN'', ``TED'', and ``AVG.'' represent the studio set, in-the-wild set, synthetic background set, temporal disturbance set and average performance across the four subsets, respectively. 
\textbf{Bold} indicates the highest value within the same data type.}
\label{mvrgb}
\setlength{\tabcolsep}{1.22em}{
\begin{tabular}{l|c|cc|cc|cc|cc|cc}
\toprule 
\multirow{2}{*}{Model} & \multirow{2}{*}{Data Type} & \multicolumn{2}{c|}{\textbf{STU}} &  \multicolumn{2}{c|}{\textbf{ITW}} & \multicolumn{2}{c|}{\textbf{SYN}} & \multicolumn{2}{c|}{\textbf{TED}} & \multicolumn{2}{c}{\textbf{AVG.}} \\
&  & Top-1 & Top-5  & Top-1 & Top-5 & Top-1 & Top-5 & Top-1 & Top-5 & Top-1 & Top-5  \\ \midrule \midrule
UMDR~\cite{UMDR} &  Pixel & \textbf{92.56} & \textbf{99.09} & 23.78& 44.22& 22.12& 42.61& \textbf{90.13}& \textbf{98.23} &57.15& 71.04   \\
KVNet-V~\cite{nla_slr} & Pixel & 91.57&99.00  & \textbf{62.25} &  \textbf{86.19} &\textbf{70.90}& \textbf{90.97} &79.78&94.68 &\textbf{76.13}&\textbf{92.71} \\
\midrule
SPOTER~\cite{SPOTER} & 2D pose &  76.92&95.55&67.79&89.98&69.21&92.16&74.34&\textbf{94.14}&72.06&92.96      \\ 
DSTA-SLR~\cite{DSTA-SLR}& 2D pose& \textbf{91.68}&	97.22&	\textbf{87.06}&	95.86&	85.67&	\textbf{96.34}&	\textbf{79.15}&	92.14&	\textbf{85.89}&	95.39\\
STC-SLR~\cite{SAM-SLR}& 2D pose& 90.11 &	96.28&	86.82&	94.91&	\textbf{86.09}&	96.29&	75.13&	90.76&	84.53&	94.56\\
\textbf{}KVNet-K~\cite{nla_slr} & 2D pose & 90.45 &\textbf{98.56}  & 86.23&\textbf{97.77} & 85.73&95.47& 77.26&93.93 &84.92& \textbf{96.43}\\
\midrule
SAM-SLR~\cite{SAM-SLR} & 2D pose + Pixel & 85.85&97.68&77.36&92.88&84.26&95.69&79.92&88.10&81.85&93.59          \\
NLA-SLR~\cite{nla_slr} & 2D pose + Pixel & \textbf{94.62} & \textbf{99.31} &  \textbf{89.75}&\textbf{98.60}  & \textbf{88.94} &\textbf{96.98}&\textbf{85.19}&\textbf{96.69}&\textbf{89.63}&\textbf{97.90} \\
\bottomrule
\end{tabular}
}
\end{table}

\begin{table}[t]
\centering
\tiny
\caption{\textbf{The baseline of Multi-view RGB-D-based ISLR on MM-WLAuslan. }}
\label{mvrgbd}
\setlength{\tabcolsep}{1.075em}{
\begin{tabular}{l|c|cc|cc|cc|cc|cc}
\toprule 
\multirow{2}{*}{Model} & \multirow{2}{*}{Data Type} & \multicolumn{2}{c|}{\textbf{STU}} &  \multicolumn{2}{c|}{\textbf{ITW}} & \multicolumn{2}{c|}{\textbf{SYN}} & \multicolumn{2}{c|}{\textbf{TED}} & \multicolumn{2}{c}{\textbf{AVG.}} \\
 & & Top-1 & Top-5  & Top-1 & Top-5 & Top-1 & Top-5 & Top-1 & Top-5 & Top-1 & Top-5  \\ \midrule \midrule
UMDR~\cite{UMDR} &  Pixel + Depth &\textbf{93.25}&\textbf{99.11}&\textbf{74.98}&\textbf{92.19}&\textbf{86.14}&\textbf{96.24}&\textbf{90.42}&\textbf{97.39}&\textbf{86.20}&\textbf{96.36}    \\
KVNet-V~\cite{nla_slr} &  Pixel + Depth &87.67 & 98.22 & 66.01 &88.80  & 83.06& 95.27  & 74.23&92.28 &77.74&93.64\\
\midrule
SPOTER~\cite{SPOTER} & 3D pose &  79.91&\textbf{96.91}&73.44&91.29&\textbf{76.41}&\textbf{93.58}&76.87&94.45&76.66&94.06\\ 
ST-GCN~\cite{pose3} & 3D pose & \textbf{81.77}&95.07&\textbf{77.34}&\textbf{93.13}&76.38&92.83&\textbf{79.36}&\textbf{96.73}&\textbf{78.71}&\textbf{94.44} \\ \midrule
SAM-SLR~\cite{SAM-SLR} & 3D pose + Pixel + Depth & 89.21&98.83&80.51&94.18&83.76&96.67&85.68&93.78&84.79&95.87 \\
NLA-SLR~\cite{nla_slr} & 2D pose + Pixel + Depth & \textbf{94.43} &\textbf{99.37}&\textbf{88.95}&\textbf{98.49}&\textbf{89.52}& \textbf{97.14}&\textbf{85.13}&\textbf{96.46}&\textbf{89.51}& \textbf{97.87} \\
\bottomrule
\end{tabular}
}
\end{table}

\begin{table}[t]
\centering
\tiny
\caption{\textbf{The baseline of Cross-Camera ISLR on MM-WLAuslan.} 
\textit{``K''}, \textit{``RS''} and \textit{``K+''} represent Front Kinect-v2, Front RealSence and Left-Front + Right-Front Kinect-v2, respectively. ``STU'', ``ITW'', ``SYN'', ``TED'', and ``AVG.'' represent the studio set, in-the-wild set, synthetic background set, temporal disturbance set and average performance across the four subsets, respectively.}
\label{cc}
\begin{tabular}{l|c|c|c|cc|cc|cc|cc|cc}
\toprule 
\multirow{2}{*}{Model} & \multirow{2}{*}{Train} & \multirow{2}{*}{Test} & \multirow{2}{*}{Data Type} & \multicolumn{2}{c|}{\textbf{STU}} &  \multicolumn{2}{c|}{\textbf{ITW}} & \multicolumn{2}{c|}{\textbf{SYN}} & \multicolumn{2}{c|}{\textbf{TED}} & \multicolumn{2}{c}{\textbf{AVG.}} \\
 &  &  &  & Top-1 & Top-5  & Top-1 & Top-5 & Top-1 & Top-5 & Top-1 & Top-5 & Top-1 & Top-5  \\ \midrule \midrule
% Model & Train & Test & Data Type & Top-1 & Top-5 & Top-1 & Top-5 & Top-1 & Top-5 & Top-1 & Top-5 & Top-1 & Top-5   \\ \midrule
\multirow{5}{*}{\parbox{0.9cm}{KVNet-V\\\centering \cite{nla_slr}}} & \textit{K} & \textit{K}  & Pixel &84.51&	97.57&	39.88&	68.00&	56.56&	82.18&	70.31&	90.86&	62.82&	84.65    \\
& \textit{RS} & \textit{RS}  &  Pixel & 66.41	& 89.58	&26.82&	52.05&	41.70&	68.52&	56.52&	82.35&	47.86&	73.12 \\ \cline{2-14}
\specialrule{0em}{0.55pt}{0.55pt}
 & \textit{K} & \textit{RS} & Pixel & 53.33&	81.06&	18.88&	41.58&	32.32&	60.09&	46.05&	71.03&	37.65&	63.44\\
 & \textit{RS} & \textit{K} & Pixel &31.28	&55.3	&5.85 &	15.73&	14.35	&30.39	&25.35	&46.55	&19.21&	36.99\\
 & \textit{RS} & \textit{K+} & Pixel &  5.36&	14.45&	1.97&	6.36&	1.97&	6.39&	3.84&	11.03&	3.28&	9.56 \\ \midrule
 \multirow{5}{*}{\parbox{0.9cm}{UMDR\\\centering \cite{UMDR}}}& \textit{K} & \textit{K}  & Pixel + Depth & 91.65	&98.81&	72.52&	90.46&	83.77&	95.18&	88.35&	98.07&	84.07&	95.63      \\
 & \textit{RS} & \textit{RS}  & Pixel + Depth &  91.34&	98.64&	75.66&	92.78&	84.25&	95.83&	86.65&	97.50&	84.47&	96.19      \\
 \cline{2-14} \specialrule{0em}{0.55pt}{0.55pt}
 & \textit{K} & \textit{RS} & Pixel + Depth &  79.09&	94.67&	44.00&	67.81&	0.64&	2.33&	71.47&	90.91&	48.80&	63.93   \\
 & \textit{RS} & \textit{K} & Pixel + Depth & 71.20&	89.87&	35.08&	59.93&	46.11&	68.40&	61.05&	83.88& 53.36& 75.52   \\
 & \textit{RS} & \textit{K+} & Pixel + Depth & 11.25&	26.67&	2.45&	8.03&	3.84&	11.37&	7.88&	19.00&	6.36&	16.27   \\
\bottomrule
\end{tabular}
\end{table}

\begin{table}[t]
\centering
\tiny
\caption{\textbf{The baseline of Cross-view ISLR on MM-WLAuslan.} 
\textit{``L''}, \textit{``F''} and \textit{``R''} represent left-front, front and right-front Kinect-v2, respectively.}
\label{cv}
% Cam_ID: 1 <-> Right
% Cam_ID: 2 <-> Left
% Cam_ID: 3 <-> Front
\begin{tabular}{l|c|c|c|cc|cc|cc|cc|cc}
\toprule 
\multirow{2}{*}{Model} & \multirow{2}{*}{Train} & \multirow{2}{*}{Test} & \multirow{2}{*}{Data Type} & \multicolumn{2}{c|}{\textbf{STU}} &  \multicolumn{2}{c|}{\textbf{ITW}} & \multicolumn{2}{c|}{\textbf{SYN}} & \multicolumn{2}{c|}{\textbf{TED}} & \multicolumn{2}{c}{\textbf{AVG.}} \\
 & & & & Top-1 & Top-5  & Top-1 & Top-5 & Top-1 & Top-5 & Top-1 & Top-5 & Top-1 & Top-5  \\ \midrule \midrule
\multirow{6}{*}{\parbox{0.9cm}{KVNet-V\\\centering \cite{nla_slr}}} & \textit{F} & \textit{F}  & Pixel &84.51&	97.57&	39.88&	68.00&	56.56&	82.18&	70.31&	90.86&	62.82&	84.65    \\
 & \textit{L} & \textit{L}  & Pixel  & 80.59&	95.74&	45.17&	71.29&	57.93&	82.92&	64.73&	86.86&	62.11&	84.20    \\
  & \textit{R} & \textit{R}  & Pixel & 80.82&	95.68&	37.97&	65.94&	37.62&	64.82&	62.80&	85.85&	54.80&	78.07    \\
\cline{2-14} \specialrule{0em}{0.55pt}{0.55pt}
 & \textit{F} & \textit{L+R}  & Pixel &  23.60&	48.10&	8.70&	23.28&	9.94&	26.53&	15.90&	35.41&	14.53&	33.33   \\
 & \textit{L} & \textit{F+R}  & Pixel  & 29.18&	48.41&	12.48&	27.28&	21.84&	40.21&	19.58&	37.16&	20.77&	38.26    \\
 & \textit{R} & \textit{F+L}  & Pixel&  24.93&	44.53&	16.93&	34.15&	20.10&	39.26&	18.99&	36.33&	20.24&	38.57   \\\midrule
\multirow{6}{*}{\parbox{0.9cm}{UMDR\\\centering \cite{UMDR}}} & \textit{F} & \textit{F}  & Pixel + Depth & 91.65	&98.81&	72.52&	90.46&	83.77&	95.18&	88.35&	98.07&	84.07&	95.63 \\
& \textit{L} & \textit{L}  & Pixel + Depth  &  91.16&	98.71&	46.90&	70.90&	79.29&	92.93&	86.74&	97.23&	76.02&	89.95  \\
 & \textit{R} & \textit{R}  & Pixel + Depth & 90.95&	98.56&	13.80&	28.72&	73.92&	90.74&	85.81&	96.87&	66.12&	78.72     \\
\cline{2-14} \specialrule{0em}{0.55pt}{0.55pt}
 & \textit{F} & \textit{L+R}  & Pixel + Depth &  32.27&	55.95&	10.06&	19.83&	21.64&	41.07&	27.32&	49.02&	22.82&	41.47    \\
 & \textit{L} & \textit{F+R}  & Pixel + Depth  &  40.55&	62.42&	6.44&	14.61&	25.58&	44.83&	32.27&	53.74&	26.21&	43.90  \\
 & \textit{R} & \textit{F+L}  & Pixel + Depth &  28.82&	47.04&	6.62&	14.73&	19.74&	36.03&	24.18&	37.45&	19.84&	33.81     \\
\bottomrule
\end{tabular}

\end{table}

All single-view experiments in this section are conducted on the data captured by front Kinect-V2.

\textbf{Single-view RGB-based ISLR:} Following previous works~\cite{wlasl,msasl,popsign_asl}, we adopt this setting as a central focus of ISLR research. 
We utilize publicly available ISLR models, such as KVNet~\cite{nla_slr}, SPOTER~\cite{SPOTER}, DSTA-SLR~\cite{DSTA-SLR}, STC-SLR~\cite{STC-SLR}, SAM-SLR~\cite{SAM-SLR} and NLA-SLR~\cite{nla_slr}.
Meanwhile, we incorporate models that have demonstrated strong performance in action recognition, including I3D~\cite{i3d}, SlowFast~\cite{slowfast} and Timesformer~\cite{timesformer}.
As indicated by Table~\ref{svrgb}, pixel-based models perform well in controlled STU.
This suggests that pixel models are effective in settings with minimal noise and well-defined conditions.
Conversely, pose-based models are robust in challenging environments, like ITW and SYN, because they focus on structural rather than textural information. % 
Furthermore, NLA-SLR~\cite{nla_slr} is the SOTA model for ISLR. 
It ensembles the high-performance KVNet-V and KVNet-K models for pixel and pose data, respectively.
The model demonstrates high accuracy across all test subsets consistently. 

\textbf{Single-view RGB-D-based ISLR: }
As shown in Table~\ref{svrgbd}, the combination of pixel and depth data generally improves recognition accuracy on most methods, highlighting the benefits brought by depth data. 
% This highlights the advantages of using both types of data to enhance model performance.
However, the performance of the KVNet-V~\cite{nla_slr} model declines, indicating its insufficient processing of depth information alongside pixel data. % ineffective
In contrast, the UMDR~\cite{UMDR} model, a SOTA model for RGB-D action recognition, leads to significant performance improvements across various test subsets.
% This demonstrates the effectiveness of models designed to utilize specific combinations of data types, thereby improving recognition accuracy in diverse environments. 
Additionally, pose-based models with 3D pose data as the input also show improved performance, further supporting the benefits of integrating depth information into pose-based models.

\textbf{Multi-view RGB-based \& RGB-D-based ISLR:} In Table~\ref{mvrgb} and Table~\ref{mvrgbd}, we show performances of several RGB-based and RGB-D-based models on multi-view ISLR. 
The results highlight that using multiple views and additional modalities generally improves model performance. 
% Models like UMDR and SAM-SLR, integrating depth or 3D pose data, show consistently high results. 
Models like UMDR and SAM-SLR, incorporating depth or 3D pose data, consistently achieve better results.
This suggests these models effectively capture more comprehensive gesture information.
However, these benefits come at the cost of increased model complexity. 
The introduction of multi-view RGBD data inevitably raises the training costs of the model. 
Additionally, information redundancy in the data can potentially interfere with the model's learning process. 
For instance, the recognition accuracy of the NLA-SLR model, when trained on multi-view RGBD data, is lower compared to when it is trained solely on RGB data.
% As the number of perspectives and modalities increases, the recognition accuracy of NLA-SLR decreases. 
% We witness redundancy and feature conflicts caused by additional input features.
For future research, we focus on developing more efficient methods to optimize performance without increasing complexity for multi-view and multi-modal data.

% \textbf{Multi-view RGB-D-based: }

\textbf{Cross-camera ISLR: }
As illustrated in Table~\ref{cc}, there is a challenge in cross-camera ISLR on the MM-WL Auslan dataset.
The results show a significant decline in accuracy when models trained on one type of camera are tested on the other one. 
Although two models, \ie, KVNet-V~\cite{nla_slr} and UMDR~\cite{UMDR}, perform well with data from the same camera, their performance drops across the cameras. This highlights the substantial differences between the two cameras, emphasizing the complexity of achieving robust ISLR across varied hardware. 
The challenge of cross-camera ISLR underscores the need for developing models that can better generalize on data from various cameras.

\textbf{Cross-view ISLR: }
We report the performance of two models, \ie, KVNet-V~\cite{nla_slr} and UMDR~\cite{UMDR}, training and evaluating on the data from different Kinect-v2 views, as shown in Table~\ref{cv}. 
UMDR, incorporating depth information alongside pixel data, generally exhibits greater resilience and performance compared to KVNet-V.
Both models exhibit high accuracy under the single-view setting of our dataset, yet experience a significant drop in accuracy in the cross-view context.
This indicates that models capable of adapting to diverse visual inputs are necessary to address the challenges posed by cross-view.

% Compared with the performance on the same view, a significant performance degradation across shifting camera views highlights the challenges in the setting. 
% Notably, UMDR, incorporating depth information alongside pixel data, generally exhibits greater resilience and performance compared to KVNet-V.
% However, the recognition accuracy remains low, underscoring the necessity for models capable of adapting to diverse visual inputs.
% The results highlight the challenges in cross-view ISLR. 
% Both models perform best when trained and tested on the same camera view, with significant performance drops when testing across different angles. 
% The UMDR model, which integrates depth with pixel data, generally shows better resilience and performance in cross-view testing compared to KVNet-V. 
% This indicates that depth information aids in handling viewpoint variations more effectively, emphasizing the need for models that can adapt to different visual inputs.

%=================================
\section{Limitation and Future Work}
% 第一点： 数据量的多样性还是小？
% 第二点： 更有效的多模态融合和多视角的方法需要提出。
% 第三点： 需要更好的解决迁移问题。
% 第四点： 建议Auslan Sign Bank 本身存在的问题，我们将设计一个双向的手语字典。
% 第五点： 手语生成任务。
\textbf{Limited Diversity in Data:} 
As shown in Figure~\ref{stat}(a) of the main paper, we analyze the distribution of Caucasian, African, and Asian signers within the MM-WLAuslan dataset. 
We observe that the proportion of African signers is significantly lower than that of Caucasian and Asian signers. 
Consequently, the signers in our recordings do not fully represent the demographic diversity of the Auslan community. 
Australia, being a multi-cultural nation, encompasses a wide range of ethnicities, and the representation of these ethnicities in our dataset is crucial. 
Therefore, we will continue recording to achieve a more balanced representation.
% The diversity of our dataset is currently limited, particularly in terms of the variety of signers. 
% The signers in our recordings do not fully represent the demographic diversity of the Auslan community. 
% We plan to continue recording to achieve a more balanced representation. 

\textbf{Lack of Real-world Scenarios:} Although we attempt to simulate real-life environments by altering backgrounds and capturing some data ``in the wild'', these settings still fall short of fully representing the complexities of real-world scenarios, such as multi-person interactions and intricate backgrounds. 
Moving forward, we intend to capture real-world Auslan glosses for a more authentic dataset.
This initiative aims to more accurately reflect the dynamic and diverse contexts in which Auslan is naturally used, thereby improving the relevance and applicability of the dataset.

\textbf{Existing Model Limitations:} In this work, we utilize publicly available deep learning models, some of which are not specifically designed for sign language. 
Consequently, developing more effective multi-modal fusion and multi-view techniques tailored to the unique characteristics of our dataset is essential.
This approach will enhance the accuracy and applicability of the models, ensuring they are better suited to address the specific challenges and nuances of isolated sign language recognition.

\textbf{Investigating Isolated Sign Language Production Task:} Sign Language Production is currently a popular task, involving not only the generation of isolated glosses~\cite{islp} but also continuous sign language~\cite{slp1,SLP2}. 
Unlike previous datasets, ours incorporates multi-view and multi-modal capabilities, enabling the creation of more accurate 2D or 3D sign language representations. 
We plan to further explore this task using our dataset in future research. 
This will enhance the precision and effectiveness of sign language modelling, providing more robust tools for communication within the deaf and hard-of-hearing community.

\section{Conclusion}
In this work, we introduce the first large-scale, multi-view, multi-modal word-level dataset for Australian Sign Language (Auslan), named MM-WLAuslan. The dataset includes 282K+ videos encompassing 3,215 distinct Auslan glosses performed by 73 signers.
To the best of our knowledge, MM-WLAuslan has the largest amount of data, the most extensive vocabulary, and the most diverse set of multi-modal camera views.
We position four RGB-D cameras, \ie, three Kinect-V2 cameras and a RealSense camera, hemispherically around the signers.
Extensive experiments demonstrate the validity and challenges of MM-WLAuslan. Thanks to the cross-camera, multi-view, and multi-modal gloss videos, our dataset can be used for practical applications related with Auslan. Furthermore, the presented benchmark results can act as strong baselines for future research.

% propose the first publicly available large-scale word-level Auslan dataset with multi-view and multi-modality, named MM-WLAuslan. 
% To the best of our knowledge, MM-WLAuslan represents the most extensive publicly available Auslan dataset in terms of vocabulary breadth and sample number. 
% Furthermore, it also stands as the largest dataset to date featuring comprehensive multi-view and multi-modal capabilities.
% Extensive experimental testing confirms that incorporating multiple views and depth information significantly enhances the accuracy of Isolated Sign Language Recognition.
% These findings not only offer valuable insights for future research but also pave the way for more practical applications of sign language recognition technologies.

\section*{Acknowledgement}
This research is funded in part by ARC-Discovery grant (DP220100800 to XY), ARC-DECRA grant (DE230100477 to XY) and Google Research Scholar Program.
We express our gratitude to Professor Trevor Cohn for his valuable feedback on this work. 
We also gratefully thank all the anonymous reviewers and ACs for their constructive comments.

\section*{Broader Impact}
The development of the word-level Australian Sign Language (Auslan) dataset has several impacts on technology, education, and society. 
Our proposed MM-WLAuslan, recorded using multi-view RGB-D cameras and focused on isolated Auslan glosses, brings about a wide range of positive effects:
\begin{itemize}
\item \textbf{Improveing Accuracy and Efficiency in ISLR}: The high-quality data provided by multi-view RGB-D cameras enhance the detailed capture of sign language gestures, which is crucial for developing efficient and accurate ISLR systems. 
% Such systems can recognize word-level Auslan in real-time with high precision, improving communication efficiency for the deaf and hard-of-hearing.

\item \textbf{Facilitating Social Integration for the Deaf:} 
% Improved ISLR technology can better support communication for the deaf and hard-of-hearing, reducing misunderstandings and communication barriers, thereby helping them integrate more effectively into society.
Improved by our MM-WLAuslan dataset, the ISLR technology can provide the deaf and hard-of-hearing community with more efficient communication capabilities. 
% It reduces misunderstandings and communication barriers, thereby facilitating the social integration of deaf and hard-of-hearing communities with hearing communities more effectively.

\item \textbf{Expanding Educational Resources:} 
% The Auslan dataset can be used as a teaching tool, helping more people learn and master Australian Sign Language. This benefits not only deaf students but also allows non-deaf individuals to learn this important communication skill. 
Our dataset can support Auslan education~\cite{sheng2024ai, DBLP:conf/acl/XuLLSS22}. By providing multi-view demonstrations, the dataset allows Auslan learners to observe signs from different views, enhancing their understanding and accuracy in sign language. 
% This immersive, comprehensive learning experience benefits not only deaf students but also non-deaf individuals, allowing them to grasp the nuances of this important communication skill more effectively and efficiently.

\item \textbf{Driving Technological Innovation:} Our dataset offers valuable resources for research in computer vision and machine learning, promoting technological development and innovation in these fields~\cite{tspnet, DBLP:journals/ijon/WeiYCSWSY23, DBLP:conf/siggrapha/ZelenskayaWLVK23, xinsmm}.
% Moreover, the multi-view multi-modal data supports interdisciplinary research, driving discoveries and technological advancements across computer science, linguistics, cognitive science, and other fields.

\item \textbf{Preserving and Promoting Culture:} By recording and utilizing the MM-WLAuslan dataset, we preserve and disseminate the unique cultural heritage of Australian Sign Language, enhancing public awareness of its cultural value~\cite{Auslan_Corpus_new}.
\end{itemize}

These societal impacts demonstrate that the development and application of the Auslan dataset are not only technically significant but also have profound positive values on social and cultural levels.

\clearpage

\clearpage
{\small
\bibliographystyle{unsrt}
\bibliography{ref}
}

\clearpage
\section*{Checklist}

%%% BEGIN INSTRUCTIONS %%%
The checklist follows the references.  Please
read the checklist guidelines carefully for information on how to answer these
questions.  For each question, change the default \answerTODO{} to \answerYes{},
\answerNo{}, or \answerNA{}.  You are strongly encouraged to include a {\bf
justification to your answer}, either by referencing the appropriate section of
your paper or providing a brief inline description.  For example:
\begin{itemize}
  \item Did you include the license to the code and datasets? \answerYes{See Section~\ref{gen_inst}.}
  \item Did you include the license to the code and datasets? \answerNo{The code and the data are proprietary.}
  \item Did you include the license to the code and datasets? \answerNA{}
\end{itemize}
Please do not modify the questions and only use the provided macros for your
answers.  Note that the Checklist section does not count towards the page
limit.  In your paper, please delete this instructions block and only keep the
Checklist section heading above along with the questions/answers below.
%%% END INSTRUCTIONS %%%

\begin{enumerate}

\item For all authors...
\begin{enumerate}
  \item Do the main claims made in the abstract and introduction accurately reflect the paper's contributions and scope?
    \answerYes
  \item Did you describe the limitations of your work?
    \answerYes{Section 5.}
  \item Did you discuss any potential negative societal impacts of your work?
    \answerNo{Our work does not pose any negative societal impacts.}
  \item Have you read the ethics review guidelines and ensured that your paper conforms to them?
    \answerYes
\end{enumerate}

\item If you are including theoretical results...
\begin{enumerate}
  \item Did you state the full set of assumptions of all theoretical results?
    \answerNo{We do not have theoretical results.}
	\item Did you include complete proofs of all theoretical results?
    \answerNo{We do not have theoretical results.}
\end{enumerate}

\item If you ran experiments (e.g. for benchmarks)...
\begin{enumerate}
  \item Did you include the code, data, and instructions needed to reproduce the main experimental results (either in the supplemental material or as a URL)?
    \answerYes{All datasets and benchmarks are available at~\href{https://uq-cvlab.github.io/MM-WLAuslan-Dataset/}{\faGithub~\textcolor{deeppurple}{MM-WLAuslan}}.}
  \item Did you specify all the training details (e.g., data splits, hyperparameters, how they were chosen)?
    \answerYes{See Section~\ref{sec:dataset}.}
	\item Did you report error bars (e.g., with respect to the random seed after running experiments multiple times)?
    \answerNo{}
	\item Did you include the total amount of compute and the type of resources used (e.g., type of GPUs, internal cluster, or cloud provider)?
    \answerYes{Appendix Section C.}
\end{enumerate}

\item If you are using existing assets (e.g., code, data, models) or curating/releasing new assets...
\begin{enumerate}
  \item If your work uses existing assets, did you cite the creators?
    \answerYes{We cite the papers of the model.}
  \item Did you mention the license of the assets?
    \answerNo{}
  \item Did you include any new assets either in the supplemental material or as a URL?
    \answerYes{}
  \item Did you discuss whether and how consent was obtained from people whose data you're using/curating?
    \answerYes{}
  \item Did you discuss whether the data you are using/curating contains personally identifiable information or offensive content?
    \answerNo{}
\end{enumerate}

\item If you used crowdsourcing or conducted research with human subjects...
\begin{enumerate}
  \item Did you include the full text of instructions given to participants and screenshots, if applicable?
    \answerNo{}
  \item Did you describe any potential participant risks, with links to Institutional Review Board (IRB) approvals, if applicable?
    \answerNo{}
  \item Did you include the estimated hourly wage paid to participants and the total amount spent on participant compensation?
    \answerYes{}
\end{enumerate}

\end{enumerate}

\end{document}